\documentclass[journal]{IEEEtran}
\usepackage{graphicx}
\usepackage{siunitx}
\usepackage{amssymb}
\usepackage[utf8x]{inputenc}
\usepackage{amsmath}
\usepackage{mathtools}
\usepackage{latexsym}
\usepackage{array}
\usepackage{url}
\DeclarePairedDelimiter\abs{\lvert}{\rvert}%
\newcolumntype{C}[1]{>{\centering\arraybackslash}m{#1}}

\usepackage[caption=false, ...]{subfig}
\usepackage{comment}
\usepackage{url}
\usepackage{xcolor}
\definecolor{newcolor}{rgb}{.8,.349,.1}
\usepackage{float}
\usepackage{xcolor}
\usepackage[linesnumbered,ruled,vlined]{algorithm2e}

\usepackage{silence}
\ifCLASSINFOpdf
\else
\fi
\begin{document}
%
\title{A Novel Approach for Partial Fingerprint Identification to Mitigate MasterPrint Generation}
%
%
%

\author{Mahesh Joshi,
        Bodhisatwa Mazumdar,
        and Somnath Dey
\thanks{Mahesh Joshi, Bodhisatwa Mazumdar, and Somnath Dey are with the Discipline of Computer Science \& Engineering, Indian Institute of Technology Indore, Khandwa Road Simrol, (Madhya Pradesh)- 453552, India. e-mail: (phd1701101004@iiti.ac.in, bodhisatwa@iiti.ac.in, somnathd@iiti.ac.in).}
\thanks{Manuscript received Month Date, Year; revised Month Date, Year.}}

%
%

\markboth{IEEE TRANSACTIONS ON INFORMATION FORENSICS AND SECURITY}%
{Joshi \MakeLowercase{\textit{et al.}}: Bare Demo of IEEEtran.cls for IEEE Journals}
%



\maketitle

\begin{abstract}
Partial fingerprint recognition is a method to recognize an individual when the sensor size has a small form factor to accept a full fingerprint. It is also used in forensic research to identify the partial fingerprints collected from the crime scenes. But the distinguishing features in the partial fingerprint are relatively low due to small fingerprint captured by the sensor. Hence, the uniqueness of a partial fingerprint cannot be guaranteed, leading to a possibility that a single partial fingerprint may identify multiple subjects. A MasterPrint is a partial fingerprint that identifies at least $4\%$ different individuals from the enrolled template database. A fingerprint identification system with such a flaw can play a significant role in convicting an innocent in a criminal case. We propose a partial fingerprint identification approach that aims to mitigate MasterPrint generation. The proposed method, when applied to partial fingerprint dataset cropped from standard FVC 2002 DB1(A) dataset, showed significant improvement in reducing the count of MasterPrints. The experimental result demonstrates improved results on other parameters, such as True match Rate (TMR) and Equal Error Rate (EER), generally used to evaluate the performance of a fingerprint biometric system.
\end{abstract}


%
\IEEEpeerreviewmaketitle

\section{Introduction}\label{sec1}
Fingerprint recognition system uses features extracted from a user's fingerprint to conclude if he is already enrolled or not. The input to such a system can be a full fingerprint or a small portion of the user fingerprint, i.e., a partial fingerprint. These systems can thus act as a security mechanism for automatic authentication and identification of an individual. In the \textit{identification} process, we assume that the user is already enrolled, but his identity is unknown. We perform a comparison of the user's fingerprint with all the records stored in the database during enrollment. The system declares the anonymous user with the identity of the record entry having the highest similarity score. For a biometric \textit{authentication} system, the user reveals his identity while submitting the fingerprint called a probe fingerprint. The system, in this case, compares the submitted fingerprint against the records of the claimed user and declare if he's what he claimed or he's lying. 

A typical fingerprint \textit{recognition} system (an identification or an authentication system) compares a probe fingerprint with a fingerprint template (secured and protected format of a biometric sample) retrieved from the database and generates a similarity score (usually a numeric value) \cite{6374609}. A user recognition is said to be successful if the similarity score is above the predefined threshold set for the biometric system. Generally, the partial fingerprint recognition system performs comparison in three possible ways. The partial probe fingerprint can undergo comparison against another partial fingerprint template, or a full fingerprint template generated directly from a full fingerprint, or a full fingerprint template created by applying image fusion technique on multiple partial fingerprints.

A fingerprint is composed of ridges, which are the dark lines and valley, which is the white portion between dark lines \cite{10.1007/0-387-23152-8_27}. Most of the fingerprint recognition system uses minutiae, a point where three ridges emerge, or a single ridge ends, as the basis for recognising an individual. The \textit{global} features in a fingerprint include core, delta, singular points, etc. \cite{DBLP:journals/tip/GuZY06}. The \textit{local} features such as ridge properties, nearest neighbours to a minutia, etc., along with the global features, play a significant role in improving the recognition accuracy of the biometric system. Due to the comparatively small dimension of a partial fingerprint, the number of minutiae extracted from it is less. So the probability of observing global features in every partial fingerprint is very low. Hence, we can find various approaches for partial fingerprint recognition in the literature, which mainly rely on extraction of minutiae and different combinations of local features.

The MasterPrint vulnerability associated with partial fingerprint identification system addressed in this paper may generate a source to mount a wolf attack. A wolf attack is a kind of presentation attack wherein a partial fingerprint acts like a wolf and identifies multiple users from the database, as shown in Fig. \ref{fig1}. The MasterPrints generated from partial fingerprints may be misused to convict an innocent in a criminal offence. The adversary can use a similar MasterPrint to break the biometric home security application or to perform an illegal transaction at the bank or money dispenser machine.

\section{The MasterPrint vulnerability}
\begin{figure}[!t]
	\begin{center}
		\includegraphics[width=8.75cm]{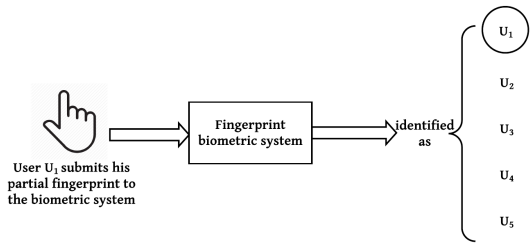}
		\caption{The MasterPrint scenario (assume there are $100$ subjects enrolled, namely, $U_{1}, U_{2}, \ldots, U_{100}$}
		\label{fig1}
	\end{center}
\end{figure}

A single (full and partial) fingerprint is generally enrolled several times to improve the accuracy of the fingerprint biometric system. We collect multiple samples of the same full fingerprint, but for recognising a partial fingerprint, the system may require more than ten partial fingerprints of the same enrolled full fingerprint. The fingerprint biometric identification system responds with a positive match if the partial probe fingerprint generates a similarity score higher than the threshold for any of the enrolled partial fingerprints. This scenario can lead to a situation where multiple partial fingerprints from different users pass the comparison test. Roy et al. \cite{DBLP:journals/tifs/RoyMR17} investigated this vulnerability for partial fingerprint identification system and claimed to be able to generate MasterPrints. A \textit{MasterPrint} is a partial fingerprint that can identify more than $4\%$ different users from the database. Fig. \ref{fig1} demonstrates an example of MasterPrint scenario. A probe partial fingerprint belonging to user $U_{1}$ is expected to identify himself. But it produces a similarity score above threshold for partial templates for user $U_{2}$, $U_{3}$, $U_{4}$, and $U_{5}$ and thus becomes a \textit{MasterPrint}.

The authors conducted the experiments on a $150$ x $150$ pixel partial fingerprint dataset cropped from full fingerprint FVC 2002 DB1 - an optical dataset. The FingerPass DB7 capacitive dataset comprises partial fingerprints of size $144$ x $144$ pixels \cite{Jia}. A partial fingerprint was considered for the comparison if it contains at least ten minutiae. Among them, MasterPrints were those partial fingerprints which were able to identify $4\%$ or more other subjects. A commercial fingerprint identification software VeriFinger $6.1$ SDK was used to generate MasterPrints. The number of partial fingerprints enrolled during the experiment was $8220$, and the results produced $1203$ MasterPrints from an optical dataset. The authors demonstrated that an adversary can use these MasterPrints to mount a dictionary attack. The experimental results conclude that a dictionary of top five MasterPrints was able to identify $26.46\%$ of users from the capacitive dataset and $65.2\%$ users from the optical dataset. Since the optical dataset is cropped from the existing full fingerprint dataset (with the constraint of $50\%$ overlap) while the capacitive dataset is already partial, the percentage of users identified with top five MasterPrints from the optical dataset is high compared to the capacitive dataset.

The multiple impressions of the same finger may vary due to pressure applied while touching the sensor, finger orientation, moisture around the sensing device, and sweating on the finger. Moreover, digital image processing techniques such as binarization, thinning may change the minutiae position by a few pixels or the minutiae orientation by some degrees. So, generally, an approximately similar feature between two fingers is allowed during recognition. Allowing nearly identical features may also increase the probability of the existence of a MasterPrint. From the authors' perspective, the enrolment of the same finger several times and accepting a query finger even if it is similar to any of the stored templates leads to MasterPrint generation. 

The development of a better partial fingerprint matcher is suggested to address the vulnerability of MasterPrints. In this paper, we propose a novel partial fingerprint identification approach that primarily targets to mitigate MasterPrint vulnerability. In our proposed method, the number of exact feature matching between probe and gallery partial fingerprint templates is more than the approximate matching. To the best of our knowledge, we are the first to address the MasterPrint vulnerability. 
%
%
%
%

\begin{figure}[!b]
	\centering
	\includegraphics[width=8cm]{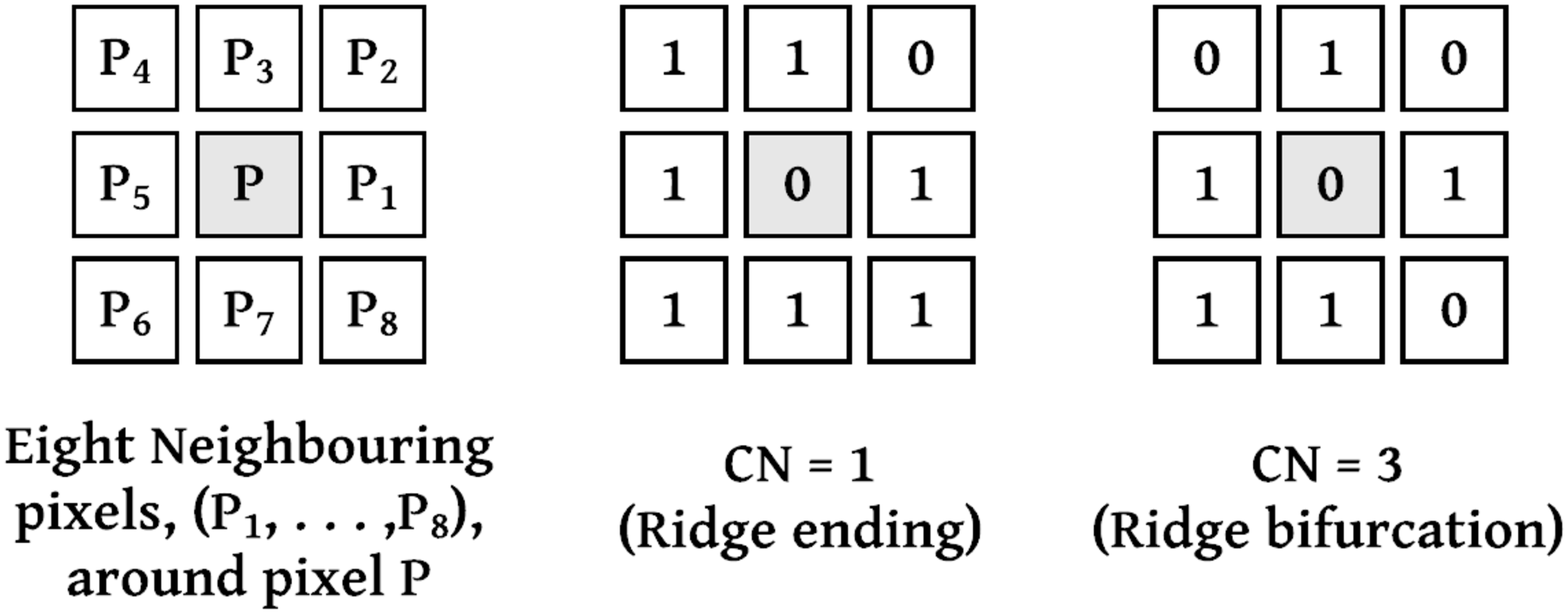}
	\caption{Significance of Crossing Number (CN) to identify a pixel as a minutia. A minutia is either ridge ending or ridge bifurcation.}
	\label{figCN}
\end{figure}

\begin{figure*}[!t]
	\centering
	\captionsetup{justification=centering}
	\includegraphics[width=17cm]{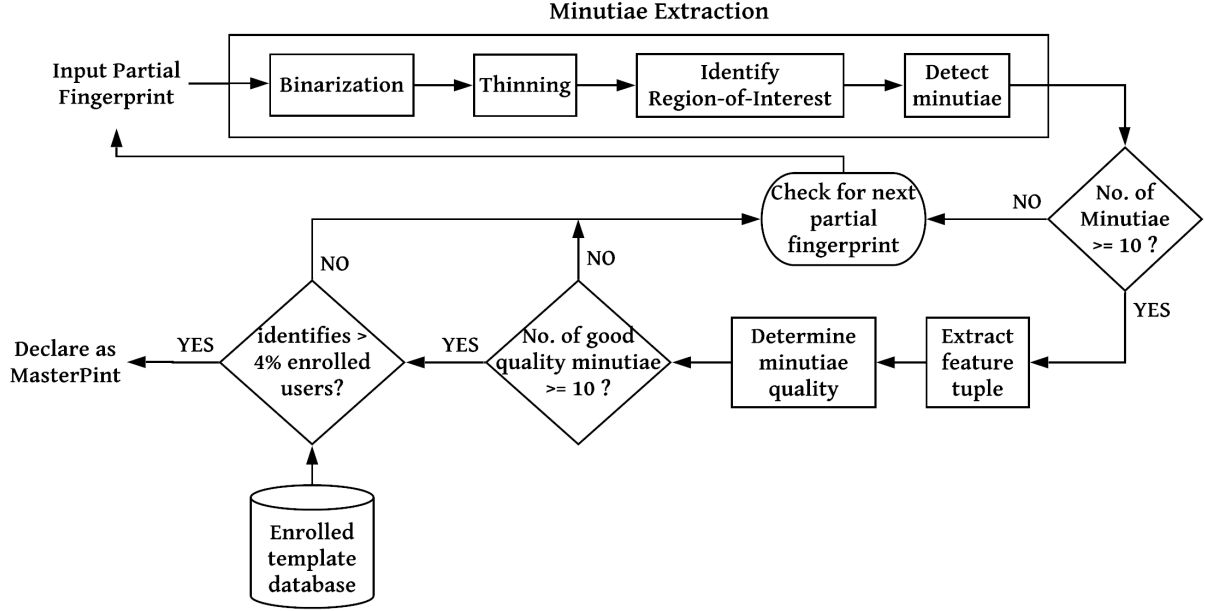}
	\caption{Flowchart for the proposed approach (assuming that the enrolment of partial fingerprints is already carried out with the system)}
	\label{fig11}
\end{figure*}

\section{Proposed approach}
Fig. \ref{fig11} shows the flowchart for the proposed approach for the partial fingerprint identification (assuming that the enrolment of all partial fingerprints from the dataset is already carried out with the system). We can roughly divide the proposed method into three steps, namely, minutiae extraction from the partial probe (to be recognised) fingerprint, extracting features and deciding minutiae quality, and MasterPrint identification. In the first step, we read a partial fingerprint, perform binarization and thinning operations, get the region-of-interest (ROI). The following subsections discusses the further steps starting from minutia detection in detail.

\subsection{True minutiae detection}

We used the metric \textit{Crossing Number} ($CN$) discussed in \cite{Rutovitz, DBLP:journals/pami/ArcelliB85, TamuraH,   DBLP:journals/mva/Mehtre93} to detect minutiae from the thinned partial fingerprint image. Fig. \ref{figCN} shows the eight-neighbourhood pixel positions, $(P_{i}, i \in \{1,\ldots,8\})$, to compute the $CN$ for the centred pixel $P$. The formula in equation (\ref{eq:cn}) computes the $CN_{p}$ for the pixel $P$. As shown in Fig. \ref{figCN}, we identify a pixel, $P$, as minutia if it's $CN$ is $1$ or $3$.

\begin{equation}\label{eq:cn}
CN_{p} = 0.5\, \cdot \left ( \left ( \sum_{i=1}^{7}\left | P_{i} - P_{i+1} \right | \right ) + \left | P_{8} - P_{1}\right | \right )
\end{equation}

Since we cropped the full fingerprint dataset to obtain partial fingerprint dataset, the pixels located near the image boundary may appear as \textit{false} minutiae (as a ridge ending). Also, the image enhancement (binarization technique) or error due to thinning operation can introduce additional \textit{false} minutiae. Therefore it is highly imperative to get rid of such spurious minutiae to improve the recognition rate. We employ Kim et al. \cite{DBLP:conf/avbpa/KimLK01} algorithm to detect and remove such false minutiae. The algorithm performs post-processing on the detected minutiae using ridge flow, ridge orientation, connectivity, and distance between minutiae. The algorithm detects and eliminates five different types of false minutiae, namely broken ridge, bridge, short ridge, hole, and triangle.

\subsection{Feature extraction}
We detect a minutia and store it's details as a triplet, $(x,y,\theta)$, where $x$ and $y$ are the position (coordinates) of the minutia and $\theta$ gives it's orientation $[\ang{0} - \ang{359}]$. We then compute the Euclidean distance of each minutia to every other minutia to find it's nearest three neighbours (by identifying lowest three distances). For a partial fingerprint with $n$ minutiae ($n>=10$), the Euclidean distance, $d_{i}, i \in \{1,2,\ldots,n-1\}$, from a minutia $m_{k}(x_{k},y_{k})$ is calculated using equation (\ref{eq:nn}). 

\begin{equation}\label{eq:nn}
d_{i} = \sqrt{(x_{k}-x_{i})^{2}+(y_{k}-y_{i})^{2}}, \,\,\,\, i \in \{1,2,\ldots,n-1\}
\end{equation}

\begin{figure}[!b]
\centering
\includegraphics[width=6cm]{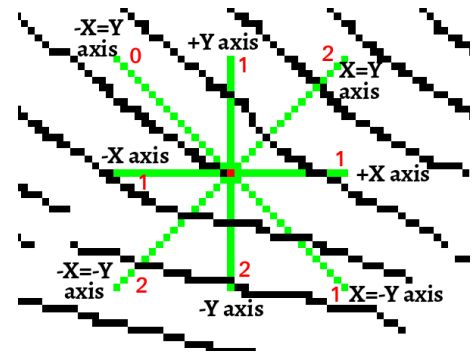}
\caption{The black squares in the figure constitute the ridge structure of a thinned partial fingerprint. The green sqaures represents the ($18$ pixel) axis formed around each minutia (marked in red color) to count the ridge crossings, $R_{cr_{i}}$, $i \in \{1,2,\ldots,8\}$, in each direction. The numbers in red colour shows the value for ridge crossings on corresponding axis.}
\label{fig4}
\end{figure}

In the next step, we count the number of ridges crossing over eight axes emerging from a minutia, as shown in Fig. \ref{fig4}. The work of Peralta et al. \cite{DBLP:journals/isci/PeraltaGTPGBBBH15} is used as a basis to empirically decide the size for all eight axes surrounding a minutia as $18$ pixels. 

Finally, we generate a $12$-element tuple corresponding to each real minutia, $M$, as

$M = (M_{q}, R_{cr_{1}}, R_{cr_{2}},\ldots,R_{cr_{8}}, d_{1}, d_{2}, d_{3})$

where, $M_{q}$ represents the minutia quality (computed using equation (\ref{eq:mq})) (0-bad quality, 1-good quality), $R_{cr_{i}}$, $ i \in \{1,2,\ldots,8\}$, is the count of ridge crossings on $i^{th}$ axis emerging from the minutia, $d_{i}$, $ i \in \{1,2,3\}$ is the Euclidean distance with the nearest three neighbours for the minutia. The values for ridge crossings, $R_{cr_{i}}$ $ i \in \{1,2,\ldots,8\}$, in a minutia tuple represents the count of the number of ridges crossing over the line formed by connecting $18$ pixels along the axis, starting from the minutia coordinates. Since the generation of similarity score requires all the entries for ridge crossing to be matched exactly, the entry for a given $R_{cr_{i}}$, $ i \in \{1,2,\ldots,8\}$, is marked with $-1$ if there is an image boundary before the $18^{th}$ pixel on its $i^{th}$ axis.

The equation (\ref{eq:mq}) represents the formula to calculate the value for $M_{q}$. This first element in each minutiae tuple is binary-valued and specifies the quality of a minutia. $M_{q}=1$ represents a good quality minutia (having all values $\geq 0$ for the ridge crossings along eight axis surrounding the minutia) which undergoes comparison with another good quality minutiae. $M_{q}=0$ represents a comparatively bad quality minutiae tuple that will miss the comparison with other minutiae tuples. Since the significance of $M_{q}$ is to discard the minutia that lies within the $18$ pixel from the image boundary, we consider the last eleven components of $M$ (we exclude $M_{q}$) for computing the similarity score.

\begin{equation}\label{eq:mq}
    M_{q} = 
\begin{cases}
    1\, ,& \text{if } \left(\,\,\sum\limits_{i=1}^{8} \abs{\,R_{cr_{i}}} - \sum\limits_{i=1}^{8} R_{cr_{i}}\right) = 0\\
    0\, , & \text{otherwise}
\end{cases}
\end{equation}

The order of the eight ridge crossing values, namely $R_{cr_{1}}, R_{cr_{2}}, \ldots, R_{cr_{8}}$, in each minutia tuple, $M$, is performed by selection of $\theta$ (stored as initial minutia triplet) to ensure \textit{rotation-invariant feature extraction}. Table \ref{tab5} shows the significance of $\theta$ to decide the first count of ridge crossings, $R_{cr_{1}}$, in a given minutiae tuple. The subsequent count of ridge crossings, $R_{cr_{i}}, i \in \{2,3,\ldots,8\}$, is measured in anti-clockwise direction from the axis corresponding to $R_{cr_{1}}$.

\begin{center}
\begin{table}[!b]
\caption{\label{tab5}Relation between $\theta$ and $R_{cr_{1}}$}\vspace{0.25cm}
\centering
\begin{tabular}{|C{2.8cm}|C{4cm}|}
\hline
\textbf{Range of $\theta$} & \textbf{Starting axis for computing $R_{cr_{1}}$}  \\
\hline
$>=$\ang{338} and $<=$\ang{22} & $+X$ axis  \\
\hline
[\ang{23} - \ang{67}] & $X=Y$ axis \\
\hline
[\ang{68} - \ang{112}] & $+Y$ axis  \\
\hline
[\ang{113} - \ang{157}] & $-X=Y$ axis  \\
\hline
[\ang{158} - \ang{202}] & $-X$ axis  \\
\hline
[\ang{203} - \ang{247}] & $-X=-Y$ axis  \\
\hline
[\ang{248} - \ang{292}] & $-Y$ axis  \\
\hline
[\ang{293} - \ang{337}] & $-Y=X$ axis  \\
\hline
\end{tabular}
\end{table}
\end{center}

\subsection{Similarity score computation}

Suppose the probe template having $n$ minutiae is represented as a set, $\mathbf{S}_{p}$, and the gallery template having $m$ minutiae is represented as a set, $\mathbf{S}_{g}$.

$\mathbf{S}_{p} = \{M_{p_{1}}, M_{p{2}}, M_{p{3}},   \ldots , M_{p{n}}\}$, and

$\mathbf{S}_{g} = \{M_{g{1}}, M_{g{2}}, M_{g{3}}, \ldots , M_{g{m}}\}$

where $M_{p_{i}}, i \in \{1,2,\dots,n\} $ and $M_{g_{j}}, j \in \{1,2,\dots,m\}$ represents a minutia tuple. Let $P$ and $G$ be the probe and gallery minutia tuples under comparison, respectively. We say the minutia $P$ and $G$ are in correspondence if the following two conditions on $R_{cr_{i}}, i \in \{1,2,\ldots,8\}$ and $d_{j}, j \in \{1,2,3\}$ are satisfied,

\begin{enumerate}
	\item ${(R_{cr_{i}})}_{P}=={(R_{cr_{i}})}_{G}\, \\ \text{and} \,{(R_{cr_{i}})}_{P} >= 0\,\\ \text{and} \,{(R_{cr_{i}})}_{G} >= 0,\,  i \in \{1,2,\ldots,8\}$
	\item ${(d_{j})}_{P} - {(d_{j})}_{G} = 0 ,\,\, j \in \{1,2,3\}$
\end{enumerate}

We create another set, $\mathbf{S}_{m}$ (empty initially), that holds the set of minutiae tuples that are in correspondence. 

$\mathbf{S}_{m} =  \{M_{1}, M_{2}, M_{3}, \ldots , M_{k}\}$ 

where $k \leq min(n,m)$ and $M_{i} \in \{\mathbf{S}_{p} \cap \mathbf{S}_{g} \}$, $ i \in \{1,2,\dots,k\}$. The count for the minutiae in correspondence is given by $m_{c}$ as

$$m_{c} = \sum_{i=1}^{min\left ( p,g \right )} \mathrm{X}_{i}$$

where, $\mathrm{X}_{i} = 1$ if there exists a pair of minutiae $M_{i} \in \mathbf{S}_{p}$ and $M_{j} \in \mathbf{S}_{g}$ in correspondence, else $\mathrm{X}_{i} = 0$. The proposed similarity score, $SS$, is then computed as the average of the sum of $\alpha_{p}$ and $\alpha_{g}$ as

\begin{equation}\label{eq:ssp}
Proposed_{SS}=\frac{\alpha_{p} + \alpha_{g}}
{2}
\end{equation}

where $\alpha_{p} = m_{c} / n$ and $\alpha_{g} = m_{c} / m$.

\section{Performance evaluation}
This section presents the details about the experimental setup and performance analysis.  The proposed approach is a partial fingerprint identification and verification system capable of mitigating the generation of MasterPrints. We target unique and accurate user identification based on partial fingerprints while mitigating MasterPrint generation. 

\subsection{Experimental setup}
We cropped the partial fingerprint dataset from the standard FVC 2002 DB1-A dataset. A desktop system with 64-bit Ubuntu 18.04.2 LTS operating system having 64 GB internal memory (RAM) and  Intel\textsuperscript{\textregistered} Xeon(R) CPU E5-1620 v3 @ 3.50 GHz × 8 processor is used for the experiment. The proposed approach is programmed using the commercial computing software MATLAB R2017a. The total number of partial fingerprints in the cropped dataset with at least $10$ minutiae is 9802, whereas 9601 partial fingerprints are having at least $10$ good quality minutiae. 

The average space requirement for storing a single minutia tuple is between $40-50$ bytes. Thus if there are $15$ tuples on an average in each template and a we have $12$ partial fingerprints per subject, then for $100$ subjects with $8$ impressions per finger would require $5.5-7.5$ MB of secondary storage space. The probability of MasterPrint generation increase as the number of comparisons with other partial fingerprint grows. In our experiment, we aimed at lowering the possibility of MasterPrint generation when the comparisons are maximum. So instead of dividing the dataset into training (enrolled) and testing (query) dataset, we compared each partial fingerprint with every other partial fingerprint from the dataset, excluding itself.

\subsection{Results targeting MasterPrint}
\begin{figure}[!b]
	\centering
	\captionsetup{justification=centering}
	\includegraphics[width=9cm]{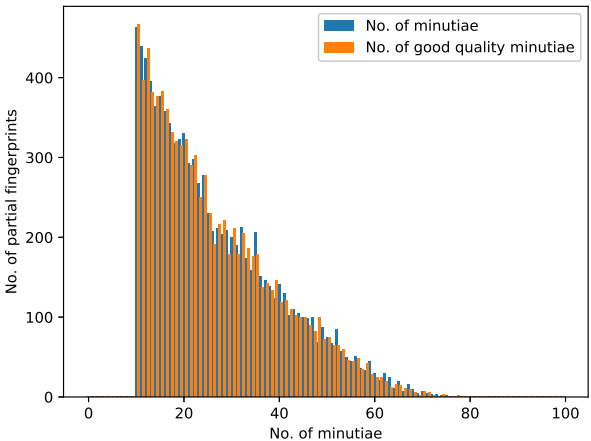}
	\caption{The analysis of number of minutiae from the dataset used in experiment.}
	\label{fig221}
\end{figure}

Fig. \ref{fig221} shows the bar chart for the number of minutiae from the dataset used in the experiments. The analysis shows that the average minutiae per partial fingerprint are 26.6, whereas the average good quality minutiae per template are 26.2. MasterPrint is an example of a closed-set biometric identification problem. Suppose, $N_{T}$ denotes the total number of comparisons, $N_{C}$ be the number of correct matches (i.e. when highest similarity score is generated for correct subject), $N_{F}$ represents the number of false matches, and $N_{R}$ be the number of rejected probe templates. The True Match Rate (TMR), False Match Rate (FMR), and False Non-match Rate (FNMR) for identification experiments are computed using the formula in equation (\ref{eq:vr2}), (\ref{eq:fr2}) and (\ref{eq:rr2}) respectively.

\begin{equation}\label{eq:vr2}
TMR = \frac{N_{C}} {N_{T}}
\end{equation}

\begin{equation}\label{eq:fr2}
FMR = \frac{N_{F}}
{N_{T}}
\end{equation}

\begin{equation}\label{eq:rr2}
FNMR = \frac{N_{R}}
{N_{T}}
\end{equation}

The Verification Rate (VR) is calculated with the formula in equation (\ref{eq:vr}).

\begin{equation}\label{eq:vr}
VR = \frac{N_{C}}
{N_{T}}
\end{equation}

Suppose we have $N$ subjects. Each subject will enrol their $J$ fingers, and there are $K$ impressions for each finger. There are L partial prints after cropping a single full fingerprint. So we have, $N_{G} = J \cdot K$, full prints per subject. Thus we have a partial fingerprint dataset, $\mathcal{F}=\{{\mathcal{PF}^{i}}_{j,k,l} | i \in \{1,\ldots,N\}, j \in \{1,\ldots,J\}, k \in \{1,\ldots,K\}, l \in \{1,\ldots,L\}\}$.

Let $\Theta$ represents the matching threshold and $S(\chi,{\mathcal{PF}^{i}}_{j,k,l})$ be the similarity score between query partial print $\chi$ and ${\mathcal{PF}^{i}}_{jkl}$. The Impostor Match Rate (IMR) is the total number of incorrect matches computed for all candidate prints when a partial fingerprint undergoes comparison with partial prints from other fingers (impostor) \cite{DBLP:journals/tifs/RoyMR17}.

\begin{equation}
IMR\left ( \chi \right ) = \frac{1}{\left ( N-1 \right )\cdot L \cdot N_{G}}\sum_{\forall i,j,k,l}\phi \left ( \chi, {\mathcal{PF}^{i}}_{j,k,l} \right )
\end{equation}

where,

$\phi \left ( \chi, {\mathcal{PF}^{i}}_{j,k,l} \right ) = \left\{\begin{matrix}
	1, & \text{if} \,\,\, S\left ( \chi, {\mathcal{PF}^{i}}_{j,k,l} \right ) > \Theta \\ 
	0, & \text{otherwise}
\end{matrix}\right.$

\begin{center}
	\begin{table}[!b]
		\caption{\label{tab6}Identification performance targeted towards zero MasterPrint generation}
		\centering
		\begin{tabular}{|C{4cm}|C{1.2cm}|C{1.3cm}|}
			\hline 
			& RSF approach & Proposed approach\\
			\hline
			Threshold & $0.0$ & $0.044$\\
			\hline
			True Match Rate (TMR) & $0\%$ & $60.37\%$\\
			\hline
			False Match Rate (FMR) & $0\%$ & $4.36\%$\\
			\hline
			False Non-match Rate (FNMR) & $0\%$ & $35.27\%$\\
			\hline
		\end{tabular}
	\end{table}
\end{center}

As the experiment was targeted to generate zero MasterPrint, the highest $IMR\left ( \chi \right )$ with a partial fingerprint is $0.315$ x $10^{-3}$ (i.e. there exists a partial fingerprint which identifies at most $3$ different subjects, including itself) proposed similarity computation approach. Table \ref{tab6} shows the analysis of the proposed approach when the target is to avoid MasterPrint generation. The TMR achieved using proposed similarity computation approach while zero MasterPrints are generated, is around $60\%$ when the False Match Rate (FMR) is $4.36\%$ and the FNMR is $35.27\%$. The TMR achieved using $SS_{1}$ while zero MasterPrints are generated, is around $61.55\%$ when the False Match Rate (FMR) is $4.37\%$ and the FNMR is $34.08\%$. 

\begin{figure}[!t]
	\centering
	\captionsetup{justification=centering}
	\includegraphics[width=9cm]{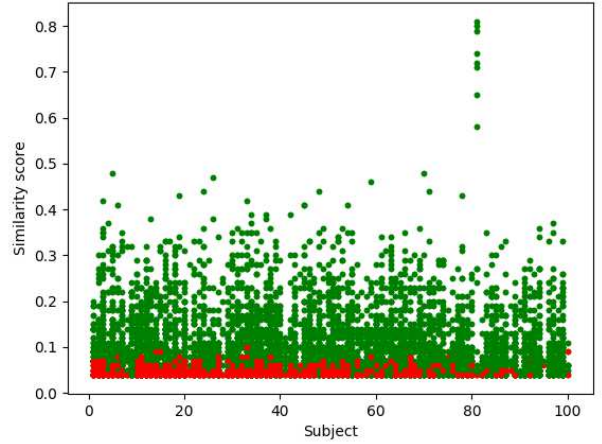}
	\caption{Scatter plot for true match and false match using similarity computation by RSF approach. The green dots represent true match where as the red dots denote false match.}
	\label{fig21}
\end{figure}

\begin{figure}[!t]
	\centering
	\captionsetup{justification=centering}
	\includegraphics[width=9cm]{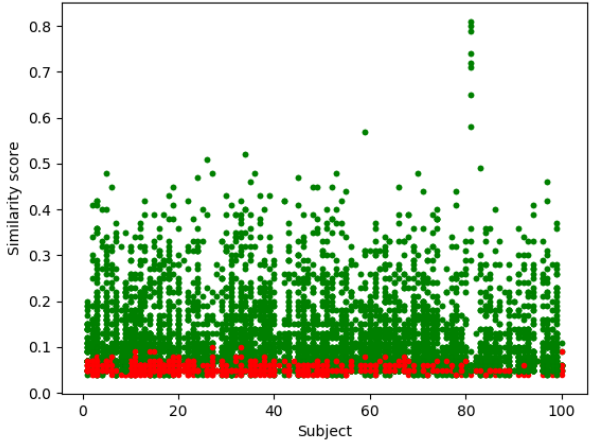}
	\caption{Scatter plot for true match and false match using proposed approach for similarity computation. The green dots represent true match where as the red dots denote false match.}
	\label{fig22}
\end{figure}

Fig. \ref{fig21} and \ref{fig22} shows the scatter plot for partial fingerprint identification having zero MasterPrint generated. The red dots show a false match, and a green dot shows a true match. Both the figures show that the false match has a low similarity score, whereas the true match generated comparatively possess high scores.

\begin{figure}[!t]
\centering
\includegraphics[width=8.5cm]{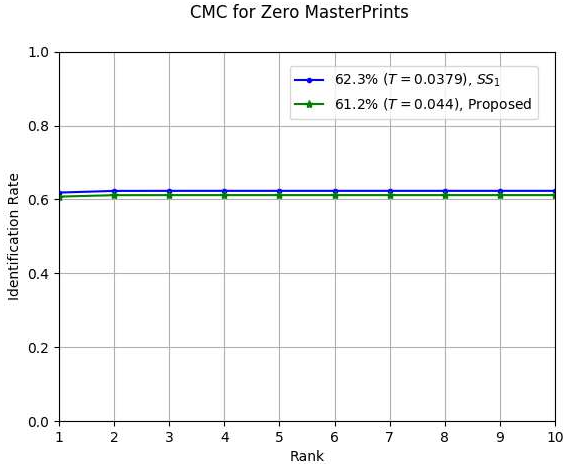}
\caption{Cumulative Matching Characteristic (CMC) curve for identifying zero MasterPrint}
\label{fig3}
\end{figure}

A Cumulative Matching Characteristic (CMC) curve shows the \textit{Rank(k)} performance of an identification system. The best performing system will have the curve more inclined towards the top-left corner of the curve. Fig. \ref{fig3} shows the \textit{Rank-10} performance of the proposed approach. The \textit{Rank-10} performance is above $61\%$. 

The CMC curve in Fig. \ref{fig3} indicates a relatively low identification rate while mitigating MasterPrint generation. The entries from Table \ref{tab6} shows that the TMR is comparatively low, and the FNMR is higher for a biometric recognition system. The typical performance achieved by a full fingerprint recognition system is usually higher than our results. After analysing these results, we concluded that two factors are affecting the identification performance of our approach. The first one is the value for $N_{T}$. We cropped a full fingerprint into $20$ partial fingerprints spanning over $4$ rows and $5$ columns as $PF_{i,j}$, where $i \in \{1,2,3,4\}$ and $j \in \{1,2,3,4,5\}$. As we have $9601$ total partial fingerprints with at least $10$ good quality minutiae, the total number of comparisons done during each experiment is $(9601 * 9600) / 2 = 46084800$. This number is very high as compared to the number of comparisons done on full fingerprints from the same dataset while identification,  $(800 * 799) / 2 = 319600$.

\begin{figure}[!t]
\centering
\includegraphics[width=8.5cm]{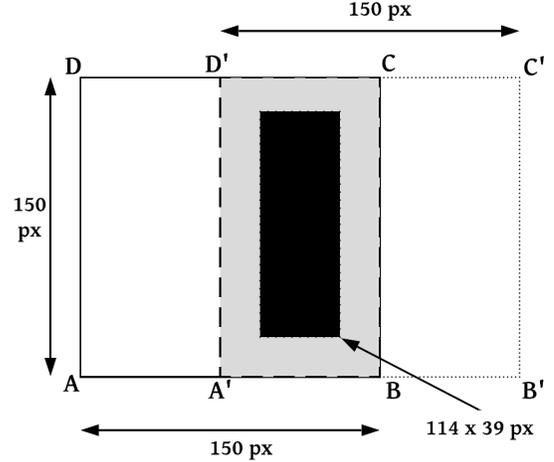}
\caption{The $50\%$ overlapping scenario for two adjacent partial fingerprints $ABCD$ and $A'B'C'D'$ cropped from a full fingerprint}
\label{fig5}
\end{figure}

The second factor is the actual region between the adjacent partial fingerprints, where the minutiae should be detected. Fig. \ref{fig5} shows this scenario. In this case, $ABCD$ and $A'B'C'D'$ are two adjacent partial fingerprints. The grey region is the $50\%$ overlapping area. Since we have $18$ pixel wide eight axes around each minutia, the minutiae from $ABCD$ and $A'B'C'D'$ observed in the black portion, and their nearest three minutiae lying in the grey region will declare a positive response after comparison. The black part is just $20\%$ of the partial fingerprint area. One can argue that we have ten full fingerprints for each subject in the dataset. Hence, we should be able to recognise from partial fingerprints cropped from these remaining full fingerprints. When we manually observed the full fingerprint dataset, we found that most of the full fingerprint images are itself a partial. 

The above two points justify the comparatively low performance of partial fingerprint identification with the dataset cropped from the standard full fingerprint dataset having $50\%$ overlap. As there is no standard partial fingerprint dataset available for experiments, any other full fingerprint dataset cropped with the similar constraint will show the performance on the lower side.

We further experimented our approach to determine the feasibility of our method towards partial fingerprint identification and verification. The observations from the result analysis are as below.

\subsection{Identification result analysis}
\begin{figure}[!b]
\centering
\includegraphics[width=8.5cm]{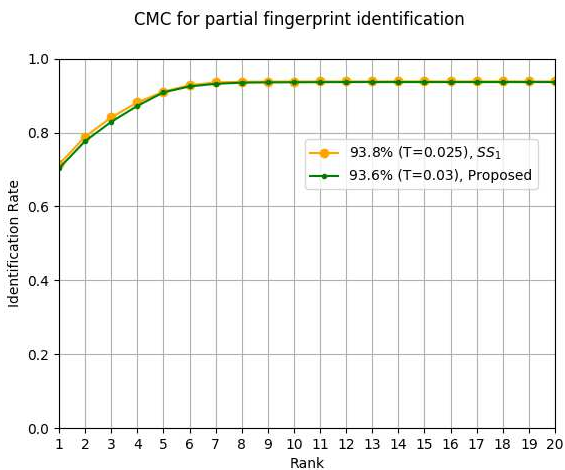}
\caption{Cumulative Matching Characteristic (CMC) curve for identifying partial fingerprints from the dataset}
\label{fig6}
\end{figure}

 The value of $\phi$ for all the results obtained for partial fingerprint identification is $0$. Fig. \ref{fig6} shows the Cumulative Matching Characteristic (CMC) curve for identifying partial fingerprints from the dataset. The curve plots the \textit{Rank-20} performance. The \textit{Rank-20} performance for partial fingerprint identification is $93.6\%$ when the threshold is $0.03$.  

Table \ref{tab7} shows the Equal Error Rate (ERR) performance of the proposed approach. The value of $\phi$ for the EER is $0$. The lower the EER, the better the biometric recognition system. As mentioned in Table \ref{tab7}, the EER achieved through the proposed approach is $0.15$. 

\begin{center}
	\begin{table}[!t]
	\caption{\label{tab7}Equal Error Rate performance}
		\centering
		\begin{tabular}{|C{3cm}|C{1.5cm}|C{1.5cm}|}
			\hline 
			 & RSF approach & Proposed approach\\
			\hline
			Threshold & $0.0$ & $0.0555$ \\
			\hline
			Equal Error Rate (EER)& $0$ & $0.15$\\
			\hline
		\end{tabular}
	\end{table}
\end{center}

The \textit{Rank-1} performance for partial fingerprint identification should be above $90\%$, and the EER should be closer to zero on an excellent performing biometric system. The justification stated above for the relatively low performance of the proposed approach while generating zero MasterPrints is also applicable here.

\subsection{Verification result analysis}

\begin{center}
	\begin{table}[!b]
		\caption{\label{tab8}Performance in verifying a subject with more than $4$ and more than $8$ partial fingerprints at $0.01\%$ FNMR}
		\centering
		\begin{tabular}{|C{4cm}|C{1.5cm}|C{1.5cm}|}
			\hline 
			 & RSF approach & Proposed approach \\
			\hline
			Threshold & $0.0$ & $0.02$ \\
			\hline
			Percentage of probes verified with $>4$ partial fingerprints & $0\%$ & $69.65\%$ \\
			\hline
			Percentage of probes verified with $>8$ partial fingerprints & $0\%$ & $39.44\%$ \\
			\hline
		\end{tabular}
	\end{table}
\end{center}

The partial fingerprint verification system requires the user to reveal his identity while submitting the partial fingerprint. The system then compares the submitted partial fingerprint against only those stored partial fingerprint with a similar identity. The zero MasterPrint generation and partial fingerprint identification results show that given a partial fingerprint, the proposed approach works well to identify an individual. This sub-section focuses on the feasibility of the proposed approach to recognise full fingerprint from the dataset.

A partial fingerprint cropped from a full fingerprint having $50\%$ overlap with the adjacent partial fingerprints will have at most $4$ other partial fingerprints that share it's $50\%$ region. Suppose we have a partial fingerprint, $p$ , that generates a similarity score above threshold for $n$ different partial fingerprints $(n>4)$. There can be at most $4$ out of $n$ partial fingerprints cropped from the same full fingerprint. The remaining $(n-4)$ partial fingerprints might have cropped from some other full fingerprint belonging to the same subject. We evaluated the proposed approach for verifying a partial fingerprint. We aimed to find the percentage of probes verified with more than $4$ and more than $8$ partial fingerprints. Table \ref{tab8} shows the verification performance of the proposed approach at Verification Rate (VR) of $99.99\%$.

The results from Table \ref{tab8} point out that a large share of probes received verification from $4$ or more partial fingerprints. We have around $70\%$ probes verified by at least five other partial fingerprints when the threshold is set to $0.02$. The result from Table \ref{tab8} gives a green signal for the possibility of the proposed approach to be able to verify a partial fingerprint cropped multiple impressions of the same finger and thus recognise a full fingerprint.

\section{Conclusion}
The MasterPrint is a vulnerability associated with partial fingerprint identification. In this scenario, a single partial fingerprint can identify at least four other distinct users enrolled with the biometric system. This paper proposed a partial fingerprint identification approach primarily target to mitigate MasterPrint generation. We detected local features associated with each minutia and performed an exact comparison for the feature set from different partial fingerprints. The results of various parameters show that the approach can generate zero MasterPrint while identifying a broad set of subjects through their partial fingerprints. The experimental results for the identification and
verification of partial fingerprints is also remarkable.


%


%

\bibliography{bibtex/bib/IEEEabrv.bib,bibtex/bib/IEEEexample.bib}{}
\bibliographystyle{IEEEtran}

%

\vskip -2\baselineskip plus -1fil

\begin{IEEEbiography}
[{\includegraphics[width=1in,height=1.55in,clip,keepaspectratio]{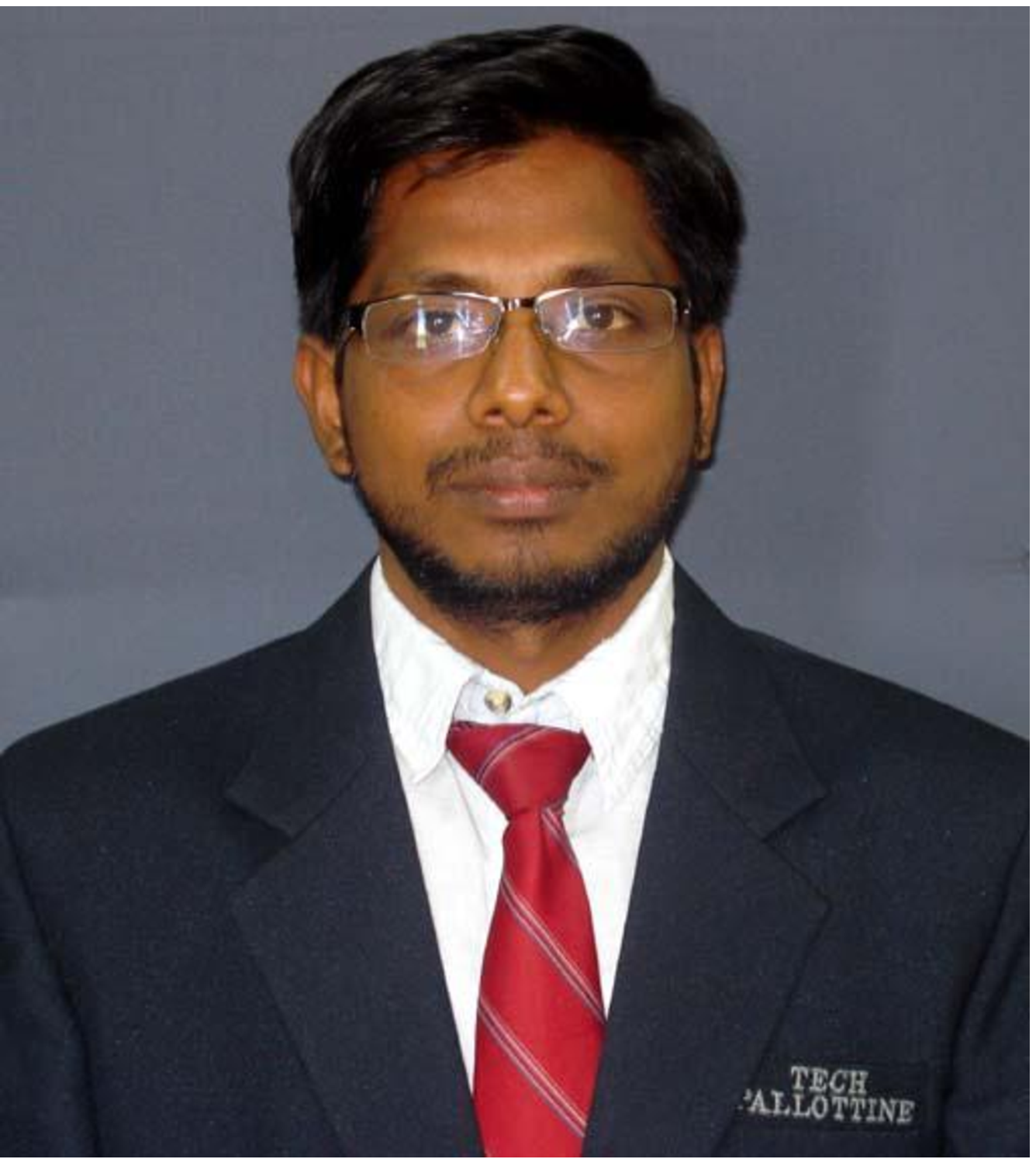}}]{Mahesh Joshi} is currently a research scholar in the Discipline of Computer Science \& Engineering at Indian Institute of
Technology Indore, India. He received M.Tech degree in
Computer Science and Engineering from the Visvesvaraya
National Institute of Technology, (formerly Visvesvaraya
Regional College of Engineering), Nagpur, India. His
current research interests include fingerprint biometrics and pattern recognition.
\end{IEEEbiography}

\vskip -2\baselineskip plus -1fil

\begin{IEEEbiography}
[{\includegraphics[width=1in,height=1.55in,clip,keepaspectratio]{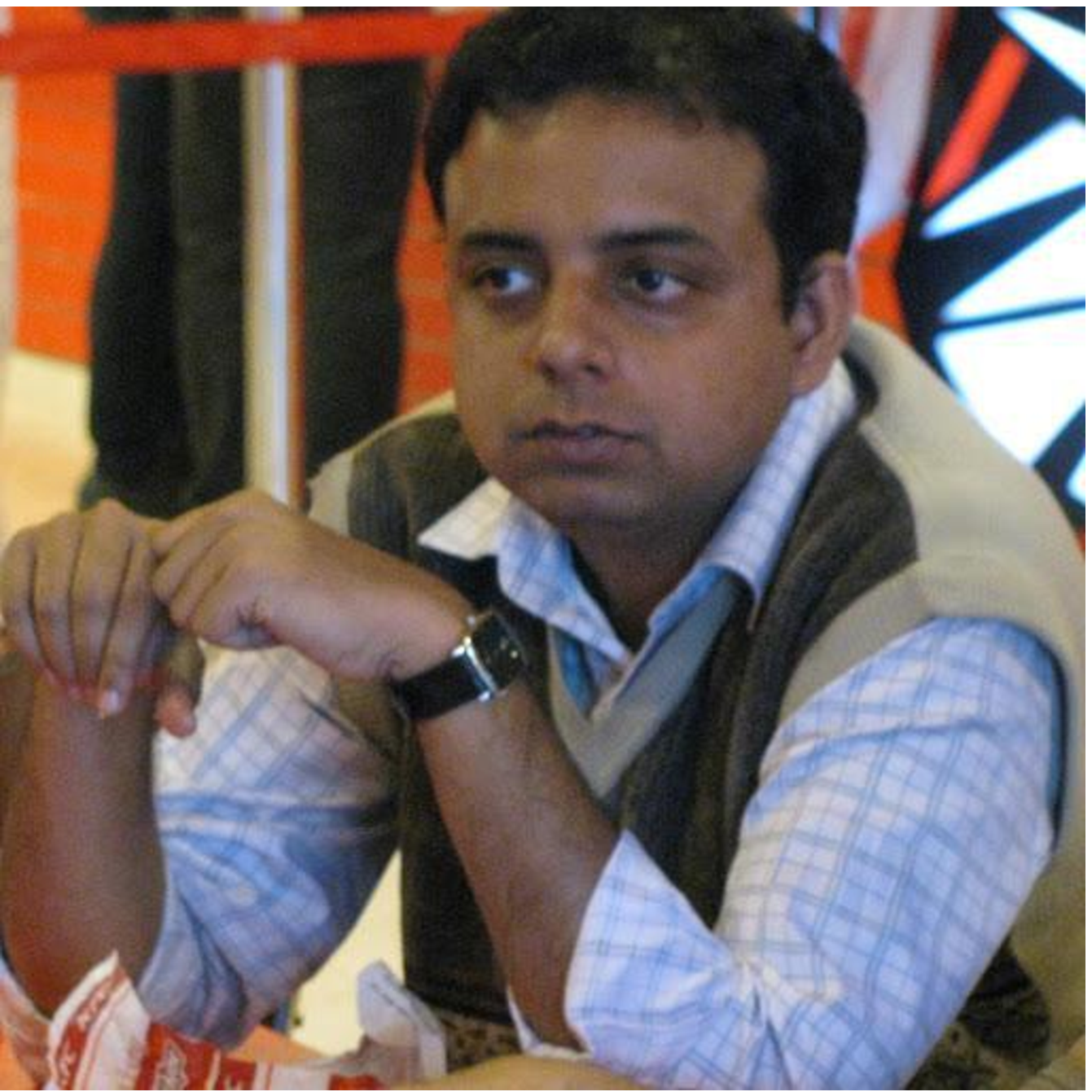}}]{Bodhisatwa Mazumdar} is currently working as an
Assistant Professor in the Discipline of Computer Science \&
Engineering at the Indian Institute of Technology Indore (IIT
Indore). He received his B.Tech degree from the University of
Kalyani and M.S. degree in electronics and electrical
communication engineering from IIT Kharagpur. He received
the PhD degree in computer science and engineering from IIT
Kharagpur. His research area is power based side-channel
analysis of cryptographic primitives like S-boxes, and security
vulnerability analysis of emerging technologies in VLSI
Design. He received the best student paper award in the 25th
IEEE International Conference on VLSI Design 2012,
Hyderabad, India. He was a post-doctoral associate in the
Design for Excellence Laboratory at New York University
Abu Dhabi, Abu Dhabi from July 2014- February 2017. He
has published over 29 research articles (including papers in
international journals, conferences).
\end{IEEEbiography}

\vskip -2\baselineskip plus -1fil

\begin{IEEEbiography}
[{\includegraphics[width=1in,height=1.55in,clip,keepaspectratio]{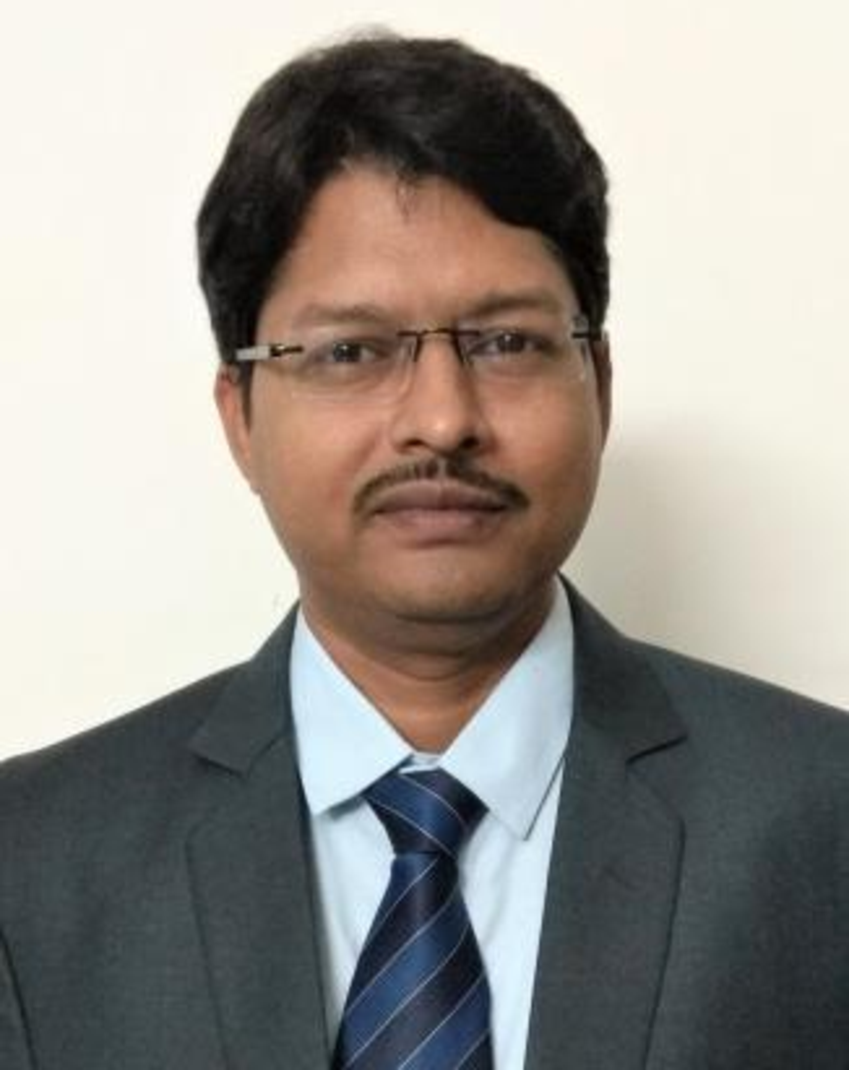}}]{Somnath Dey} is currently working as an Assistant
Professor in the Discipline of Computer Science \&
Engineering at the Indian Institute of Technology Indore (IIT
Indore). He received his B. Tech. degree in Information
Technology from the University of Kalyani in 2004. He
completed his M.S. (by research) and Ph.D. degree in
Information Technology from the School of Information
Technology, Indian Institute of Technology Kharagpur, in
2008 and 2013, respectively. His research interest includes
biometric security, biometric template protection, biometric
crypto system. He has published over 30 research articles
(including papers in international journals, conferences and
book chapters).
\end{IEEEbiography}







\end{document}